\documentclass{llncs}
\usepackage{graphicx}
\usepackage{subcaption}
\usepackage{booktabs}
\usepackage{placeins}
\usepackage{algorithm}
\usepackage{algpseudocode}

\begin{document}
\title{Layered TPOT}
\subtitle{Speeding up Tree-based Pipeline Optimization}
%
%
\author{Pieter Gijsbers\inst{1} \and Joaquin Vanschoren\inst{1} \and Randal S. Olson\inst{2}}
\institute{Technische Universiteit Eindhoven
\and University of Pennsylvania}

\maketitle
\begin{abstract}
With the demand for machine learning increasing, so does the demand for tools which make it easier to use.
Automated machine learning (AutoML) tools have been developed to address this need, such as the Tree-Based Pipeline Optimization Tool (TPOT) which uses genetic programming to build optimal pipelines.
We introduce Layered TPOT, a modification to TPOT which aims to create pipelines equally good as the original, but in significantly less time.
This approach evaluates candidate pipelines on increasingly large subsets of the data according to their fitness, using a modified evolutionary algorithm to allow for separate competition between pipelines trained on different sample sizes. 
Empirical evaluation shows that, on sufficiently large datasets, Layered TPOT indeed finds better models faster.
\end{abstract}

\section{Introduction}
The field of Automated Machine Learning (AutoML) aims to automate many of the tasks required to construct machine learning models, hence lowering the barrier to entry and yielding better models, faster. 
AutoML methods typically automate one or more steps in the creation of useful machine learning pipelines, such as the selection of preprocessing or learning algorithms, hyperparameter optimization, or a combination of them. A few methods even construct and optimize entire pipelines, such as the Tree-based Pipeline Optimization Tool (TPOT)\cite{Olson2016EvoBio}.
TPOT uses genetic programming to evolve optimal pipelines, aiming to find machine learning pipelines yielding accurate predictive models while trying to keep the pipeline as simple as possible.\\

In this paper, we introduce a novel improvement of TPOT, aimed at reducing the time needed to evaluate pipelines, without reducing the quality of the final pipeline. Indeed, the most time-consuming part in the optimization process is evaluating the performance of candidate machine learning pipelines.
In our modification, this time is reduced by initially evaluating the pipelines on a small subset of the data, and only allowing promising pipelines to be evaluated on the full dataset.
In order to do this in a fair manner, modifications to the evolutionary algorithm are implemented to prevent direct comparison between pipelines which are evaluated on different subsets of the data. As such, we aim to find pipelines of similar quality in much less time, making the tool more accessible and practical by requiring less computational time.
We call this improvement Layered TPOT (LTPOT).\\

This paper is organized as follows. First, we review related work in Sect.~\ref{sec:rw}. Then, in Sect.~\ref{sec:methods}, we discuss the proposed modification to TPOT in detail.
Next, in Sect.~\ref{sec:results}, we lay out how LTPOT will be evaluated, discuss the results of these evaluations, and propose aspects of LTPOT which can be researched in future work.
Finally, we conclude the study in Sect.~\ref{sec:conclusion}.

\section{Related Work}\label{sec:rw}
The field of AutoML is a culmination of work in the fields of algorithm selection, hyperparameter optimization and machine learning. Several AutoML systems support at least some form of 
automatic pipeline construction.\\

\begin{sloppypar}
Auto-WEKA~\cite{Thornton:2013:ACS:2487575.2487629} uses Sequential Model-based Algorithm Configuration (SMAC) to do combined algorithm selection and hyperparameter optimization~\cite{HutHooLey10-TR}, which is an adaptation of Sequential Model-Based Optimization from statistical literature. Auto-WEKA uses algorithms from the WEKA library~\cite{Hall:2009:WDM:1656274.1656278}, to build pipelines with one learner (possibly an ensemble) and optionally a single feature selection technique.\\

Auto-sklearn~\cite{NIPS2015_5872}, built on the Python library scikit-learn~\cite{Pedregosa:2011:SML:1953048.2078195}, is a re-implementation of Auto-WEKA with two extensions.
The first is the use of meta-learning to warm-start the Bayesian optimization procedure, a technique which has earlier been proven useful in \cite{Feurer:2015:IBH:2887007.2887164}.
The second addition is the automatic construction of ensembles from models evaluated during the optimization process.
Furthermore, auto-sklearn allows for more preprocessing steps to be included in the pipeline: it allows for one feature preprocessor, which includes feature extraction techniques as well as feature selection techniques, in addition to up to three data preprocessor methods, such as missing value imputation and rescaling.\\
\end{sloppypar}

TPOT differs from Auto-sklearn and Auto-WEKA by using  an evolutionary algorithm instead of SMAC. Additionally, TPOT uses a tree-based structure to represent pipelines, and considers pipelines with any number of preprocessing steps. Hence, TPOT is not constrained in the number nor the order of preprocessing steps.\\


The main idea of LTPOT is to first evaluate pipelines on a subset of the data, to get an indication of whether or not the pipeline is promising relative to other pipelines.
This idea has been explored before, for example in Sample-based Active Testing~\cite{Abdulrahman:2015:ASV:3053836.3053845}, or algorithm selection using learning curves~\cite{vanRijn2015}, where promising algorithm configurations are first evaluated on a smaller data sample to create a proxy for their performance on the full dataset. 

The use of subsets to evaluate machine learning configurations is also used in Hyperband~\cite{DBLP:journals/corr/LiJDRT16}.
Hyperband dynamically allocates resources to more promising algorithm configurations, based on experiments executed on gradually more resources.
One application that is explored is using data samples as a resource, and using increasingly bigger subsets of the data to evaluate the algorithm configurations with.
In a specific selection of datasets they showed that using this technique improved over Random Search, SMAC and Tree-structured Parzen Estimators, as introduced in \cite{NIPS2011_4443}.

In the context of evolutionary algorithms, evaluating individuals on a subset of the data to filter out bad individuals is also implemented in GTMOEP~\cite{Zutty:2016:ITE:2908961.2931641}.
Rather than a successive halving approach, they perform a selection of individuals based on evaluation on 1\% of the data.
However, there is very little evaluation done to have a clear picture of the effects of this approach on the resulting pipelines across datasets.

\section{Methods}\label{sec:methods}
In this section, we describe the modifications made to TPOT.
First, we give a brief description of how TPOT constructs and optimizes machine learning pipelines.
Then, we describe the Layered TPOT (LTPOT) structure and motivate its design.

\subsection{Structure of TPOT}

\begin{figure}
\centering
\includegraphics[scale=0.6]{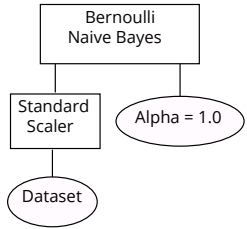}
\caption{A visual representation of a tree-based machine learning pipeline that applies standard scaling to the data before learning a Bernoulli Naive Bayes model.}
\label{fig:pipeline-tree}
\end{figure}

To construct and optimize machine learning pipelines, TPOT uses tree-based genetic programming~\cite{Banzhaf:1998:GPI:280485}.
Pipelines are represented by genetic programming trees.
An example of a tree representation of a machine learning pipeline is shown in Fig.~\ref{fig:pipeline-tree}.
The tree consists of several nodes, which can either be Primitives or Terminals, in Fig.~\ref{fig:pipeline-tree} depicted as squares and ellipses, respectively.
Primitives are operators which require input, such as an algorithm requiring data and hyperparameter values.
Terminals are constants which can be passed to primitives.
Finally, a primitive can also be used as input to a primitive, as can be seen in Fig.\ref{fig:pipeline-tree}, where the StandardScaler primitive provides the scaled dataset to the Bernoulli Naive Bayes algorithm.
Information of required input and output types is used to ensure that only valid pipelines are constructed.\\

The evolutionary algorithm then works by using these machine learning pipelines as their individuals.
It will perform mutation, for example changing a hyperparameter or adding a preprocessing step, as well as crossover, by selecting two pipelines which share a primitive, which allows them to exchange subtrees or leaves.
Finally, pipelines are evaluated and assigned a fitness score, so that a selection procedure can determine which individuals should be in the next generation.\\

As mentioned earlier, in theory these pipeline trees could be arbitrarily large.
However, very extensive machine learning pipelines are often undesirable.
With more hyperparameters, long pipelines can be harder to tune, be more prone to overfitting,  hinder understanding of the final model, and require more time to be evaluated, slowing down the optimization process.

Because of these reasons, we use a multiobjective optimization technique, NSGA-II\cite{996017}, to select individuals based on the Pareto front of the trade-off between the pipeline length and its performance.\\


\subsection{Layered TPOT Concept}
During the TPOT optimization process, most time is spent on evaluating machine learning pipelines.
Every machine learning pipeline is evaluated on the full dataset, which can take a lot of time.
Layered TPOT (LTPOT) aims to reduce the amount of pipelines evaluated on the entire dataset by doing a selection process, so that only the most promising pipelines need to be evaluated on the full dataset.
It achieves this by evaluating pipelines on a small subset of the data, and only if a pipeline exhibits good performance on that subset it will be evaluated on more data.

\subsubsection{Age-Layered Population}
To incorporate a fair evaluation of pipelines on subsets gradually increasing in size, we designed a layered structure to separate competition among individuals.
In this, we were inspired by the Age-Layered Population Structure (ALPS) \cite{Hornby:ALPS}, where individuals are segregated into layers to reduce the problem of \emph{premature convergence}.
The problem of premature convergence occurs when the individuals in the population converge to a good local optimum, which means that any new individuals are unlikely to be competitive with this local optimum, and through selection are filtered out of the population before they themselves can converge to a local optimum.
In ALPS, all individuals were given an \emph{age} which would increase as an individual or its offspring would remain in the population, and perform breeding and selection only in separate age groups called layers.
This segregation is important, because otherwise the old, locally well optimized, individuals would often prevent young individuals from being able to survive the multiple generations they needed to get closer to their local optimum.

\subsubsection{Layers in LTPOT}
In LTPOT, we wish to evaluate pipelines on subsets of the data.
However, the performance of a machine learning pipeline is influenced by the amount of training data it receives.
This means that when evaluating individuals on different subsets of the data, their performance cannot directly be compared to one another.
Therefore, the individuals are segregated into \emph{layers}.

\begin{figure}
\centering
\includegraphics[scale=0.5]{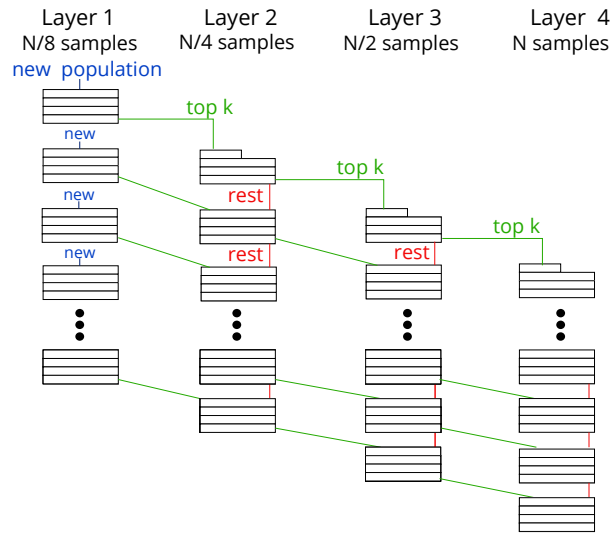}
\caption{A visual overview of the Layered TPOT structure. After a certain number of generations, each layer passes their best $k$ individuals on to the next layer, while the first layer will be provided with a newly generated set of individuals.}
\label{fig:LTPOT-vis}
\end{figure}

At each layer, individuals will be evaluated on a subset of different size.
The layers are ordered, such that in the first layer the subset used to evaluate the individuals is the smallest, and every layer above that will increase the subset size.
When an individual performs well in one layer, it will eventually be transferred to the next. This way, only the best pipelines will be evaluated on the entire dataset in the last layer.
A visual overview of the structure and flow of LTPOT is given in Fig.\ref{fig:LTPOT-vis}.\\ 

\subsubsection{Correlation of performance between layers}
In the extreme, the selection procedure implicitly assumes that the relative performance of two pipelines is the same when evaluated on a subset of the dataset as it is on the entire dataset.
However, this assumption does not always hold.
The learning curves for two pipelines may cross, meaning that one pipeline performs better after being trained on a small subset of the data, while the other performs better when trained on the full dataset.
Generally speaking, as the pipelines are evaluated on more data, the relative performance correlates more strongly with the relative performance obtained when they are trained on the entire dataset.
This is why our design will evaluate the pipelines on gradually larger subsets of data, so that when a pipeline performs worse than expected as the dataset increases, it need not be evaluated on bigger datasets.
Unfortunately, in the case where a pipeline has poor performance on a small subset, but good performance on the entire dataset, LTPOT will not pick up this pipeline.\\
We set up an experiment to verify whether or not there is indeed a correlation between the performance of a pipeline on a sample of the dataset and its performance on the entire dataset.
In this experiment, we evaluated 50 pipelines on 12 datasets with ten times 10-fold cross-validation, with various samples sizes of the dataset as well as the entire dataset.
All datasets were part of the Penn Machine Learning Benchmark (PMLB)~\cite{DBLP:journals/corr/OlsonCOUM17}.
The average AUROC of each pipeline for each sample size was determined for each dataset. Sample sizes were $\{N/2^1,\dots,N/2^5\}$, where $N$ is the number of instances in the dataset.
Because the evolutionary algorithm performs pipeline selection based on the ranking of the algorithms, rather than the value of the score metric, we ranked the averaged scores and computed the correlation of the rankings.

\addtolength{\tabcolsep}{2pt}  
\begin{table}
\centering
\begin{tabular}{lcc | rrrrrrr}
\toprule
           dataset &     instances &   features &  $\rho_{\frac{N}{2}}$ &  $\rho_{\frac{N}{4}}$ &  $\rho_{\frac{N}{8}}$ &  $\rho_{\frac{N}{16}}$ &  $\rho_{\frac{N}{32}}$ \\
\midrule
         satimage &  6435 &  36  & 0.984 & 0.903 & 0.833 & 0.759 & 0.641 \\
            clean2 &  6598 & 168  & 0.972 & 0.946 & 0.890 & 0.825 & 0.765 \\
       ann-thyroid &  7200 &  21 & 0.982 & 0.967 & 0.946 & 0.924 & 0.828 \\
           twonorm &  7400 &  20  & 0.978 & 0.929 & 0.898 & 0.759 & 0.829 \\
         mushroom &  8124 &  22 & 0.973 & 0.962 & 0.935 & 0.867 & 0.781 \\
  agaricus-lepiota &  8145 &  22  & 0.989 & 0.980 & 0.920 & 0.877 & 0.805 \\
          coil2000 &  9822 &  85  & 0.939 & 0.878 & 0.815 & 0.568 & 0.529 \\
         pendigits & 10992 &  16  & 0.984 & 0.983 & 0.945 & 0.844 & 0.771 \\
           nursery & 12958 &   8  & 0.995 & 0.995 & 0.857 & 0.917 & 0.917 \\
             magic & 19020 &  10  & 0.988 & 0.974 & 0.961 & 0.947 & 0.934 \\
            letter & 20000 &  16  & 0.990 & 0.959 & 0.918 & 0.899 & 0.801 \\
            krkopt & 28056 &   6  & 0.984 & 0.942 & 0.917 & 0.913 & 0.750 \\
\bottomrule
\end{tabular}
\caption{An overview of the spearman $\rho$ calculated by comparing the ranking of pipelines trained on subsets of the data compared to the entire dataset, $p<0.0001$ in all cases. The subscript in the column denote the size of the subset.}
\label{tab:sample_correlation}
\end{table}
\addtolength{\tabcolsep}{-2pt} 

In Table~\ref{tab:sample_correlation} the Spearman $\rho$-values are displayed, that signify the correlation between the ranking of pipelines trained on a sample of the dataset, and the ranking of pipelines when trained on the entire dataset.
The $p$-values are omitted because in all cases they are smaller than $0.0001$ and all correlations are thus significant.
From Table~\ref{tab:sample_correlation} we see that that there is a positive correlation between the rankings for all sample sizes and datasets, and the correlation gets stronger as the sample size gets closer to the full dataset size.

We experimented with various curve fitting methods to extrapolate the learning curves of pipelines so that crossing learning curves might be predicted earlier, but they did not improve the results. In future work the use of meta-learning for learning curve extrapolation, such as in~\cite{vanRijn2015}, will be tried.

\FloatBarrier 

\subsection{Layered TPOT Algorithm}
We will now give a more in-depth break down of the algorithm used in LTPOT.
Algorithm~\ref{alg:layeredea} shows the core of the layered algorithm \textproc{LayeredEA}, and Algorithm~\ref{alg:support} gives descriptions of the subroutines called from \textproc{LayeredEA}.

\begin{algorithm}
\caption{Layered Evolutionary Algorithm}\label{alg:layeredea}
\begin{algorithmic}[1]
\Function{LayeredEA}{population, S, g, G, D}
\Statex {\bf population}: a set of pipelines that will be the first generation
\Statex {\bf S}: set containing the sample size for each layer
\Statex {\bf g}: interval in generations for when a transfer should occur
\Statex {\bf G}: total number of generations
\Statex {\bf D}: dataset to construct a pipeline for
\State $M \gets \|S\|$ \Comment{Denote the number of layers.}
\State $P \gets \|population\|$ \Comment{Denote population size.}

\State $L_1 \gets$ population
\State $L_2, \dots, L_M \gets \emptyset$

\For {$i$ in $1..G$}
  \For {$l$ in $1..M$}
      \If{$L_l \neq \emptyset$ \textbf{and} $i \bmod g < 2^{(M-l+1)}$ }
          \State offspring $\gets$ \Call{VarOr}{$L_l$, $P$}
          \State \Call{Evaluate}{offspring, $S_l$, $D$}
          \State $L_l \gets$ \Call{Selection}{$L_l$ $\cup$ offspring, $P$}
          \EndIf
  \EndFor
  \If {$i \bmod g$ = 0}
  	\For {$l$ in $(M-1)..1$}
    	\State $L_{l+1} \gets L_{l+1}$ $\cup$ \Call{Top}{$L_l$, $P/2$}
    \EndFor
    \State $L_1 \gets \Call{NewPopulation}{P}$
  \EndIf
\EndFor
\State \Return \Call{Top}{$L_M$, $1$}
\EndFunction
\end{algorithmic}
\end{algorithm}

\begin{algorithm}
\caption{Functions called in Layered EA}\label{alg:support}
\begin{algorithmic}[1]
\State{{\bf function} VarOr($Population, N$)}
	\Statex Performs mutation and crossover on the individuals in $Population$, creating $N$ new individuals.
\Statex{{\bf end function}}
\Statex
\State{{\bf function} Evaluate($Population, s, D$)}
	\Statex Evaluates each individual on a subset of dataset $D$, created by taking $s$ instances by stratified sampling. Individuals are evaluated based on 3-fold CV accuracy and number of components in the pipeline. Results are saved as attributes of individuals.\footnote{Is it fine to use attributes like this in the pseudocode? It is like this in the actual code. I'm not sure if this makes it harder to understand for the reader than introducing eg. a dictionary which maps individual to fitness.}
\Statex{{\bf end function}}
\Statex
\State{{\bf function} Selection($Population, p$)}
	\Statex Creates pareto-fronts based on the accuracy-pipeline complexity trade-off. Then takes the first $p$ individuals after ordering the population by which pareto-front they are in (individuals in the first front come first). 
\Statex{{\bf end function}}
\Statex
\State{{\bf function} Top($Population, k$)}
	\Statex Returns the $k$ best individuals of the population, by constructing a pareto-front based on the accuracy-pipeline complexity trade-off.
\Statex{{\bf end function}}
\Statex
\State{{\bf function} NewPopulation($P$)}
	\Statex Creates a new population of $P$ individuals.
\Statex{{\bf end function}}
\Statex
\end{algorithmic}
\end{algorithm}

\subsubsection{Selecting parameter values}
Before calling \textproc{LayeredEA}, the number of layers as well as their \emph{sample size} is defined.
The sample size of a layer dictates how many instances are sampled from the dataset, to create the subset that the pipelines in that layer will be evaluated on.
The subset is created by stratified uniform random sampling without replacement, and pipelines will be evaluated on this subset with 5-fold cross-validation.
In this study, the sample sizes used in each layer are dictated by the size of the dataset. Let the dataset contain N instances, then the final layer will always train on the entire dataset, and each subsequent layer will use half of the data the layer above did. In this study, the number of layers used is $4$, for every dataset. The respective sample sizes used at each layer are thus $\frac{N}{8}$, $\frac{N}{4}$, $\frac{N}{2}$ and $N$. In this paper we use the term \emph{higher} layer loosely to denote layers which sample more of the entire dataset (i.e. the layer with sample size $\frac{N}{2}$ is higher than the layer with sample size $\frac{N}{4}$).

There are two ways to specify for how long the main loop of lines 7 through 20 should run: a set amount of generations $G$, or an amount of time. We chose not to work with a specific number of generations $G$, but instead let the main loop in lines 7 through 20 run for eight hours. For parameter $g$, the amount of generations between transfer, we experimented with values $2$ and $16$. With a $g$ value of $2$, LTPOT acts almost as a filter, allowing only very little optimization in early layers. A value of $16$, however, allows many generations of early optimization, before passing individuals through to the next layer.

For the final part of the initialization, a population of size $P$ is generated randomly.

\subsubsection{LayeredEA}
When calling \textproc{LayeredEA}, as shown in Algorithm~\ref{alg:layeredea}, the first step is to denote constants based on the input, specifically the number of layers $M$ (line 1) and the size of the population $P$ (line 2), as well assigning the initial population to the first layer (line 4) and marking all other layers as empty (line 5).

Then, the evolutionary algorithm will start iterating through the generations (line 6-20), at each generation again progressing the evolutionary algorithm in each layer (line 7-13) and then if required transferring individuals from one layer to the next (line 14-19).

Progressing the evolutionary algorithm in each layer (line 7-13), only happens for layers which are \emph{active}. This means that there must be a population in the layer (first clause on line 8), which it may not yet have if not enough transfers have taken place yet.
Additionally, layers which evaluate on more data are not active every generation (second clause on line 8), this is motivated below.

When progressing the evolutionary algorithm, it executes the same steps as TPOT would.
First, a new population is created from the individuals evaluated during the last generation in the same layer, by performing mutation and cross-over (line 9). However, every time a layer is passed new individuals from a previous layer, as well as in the very first generation, the provided population is taken as is without creating offspring.\footnote{This is not incorporated in the pseudo-code of algorithm~\ref{alg:layeredea}, to keep the general structure clear.}
The new individuals are then evaluated based on the sample of the data as defined by their layer (line 10).
Finally, based on the Pareto-front of the trade-off between the performance score of the pipeline as well as the pipeline length, the best individuals are picked among the new individuals as well as last generation's (line 11).
Then, every $g$ generations, the best individuals from each layer get passed to the next one. In our configuration we chose to transfer half of the layer's population.

\noindent The final pipeline chosen by LTPOT is the pipeline which has the best score in the highest layer (line 21).

\subsubsection{Optimizations}\label{subsec:opt}
There are a few scenario's that either require some additional clarification, or deviate from the above algorithm:\\

The first time a layer receives individuals from the layer before it, the selection procedure will oversample from this population so that the population in the layer will also grow to size $P$ (line 11).
This is done so that in subsequent generations, more variations of the original pipelines will exist, allowing for a better search for optimal pipelines.\\

Secondly, if LTPOT runs for a specified amount of generations, layers will be turned off whenever their population can no longer reach the highest layer.
For example, let LTPOT be configured with $4$ layers. When LTPOT is less than $3*g$ generations away from completion, any individual in the lowest layer will never reach the highest layer, thus rendering any results obtained in this layer useless.
Whenever this happens, the respective layer will no longer have their individuals evaluated or transferred to a next layer.\\

\begin{figure}
\centering
\includegraphics[scale=0.8]{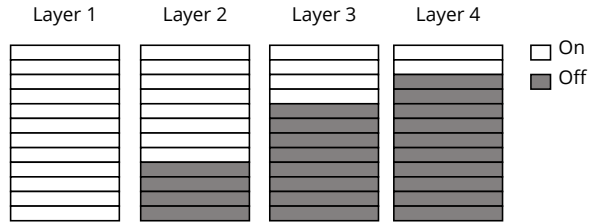}
\vspace{1cm}
\caption{Not every layer should run experiments every generation. This figure illustrates that the higher layers will be 'turned off' in higher layers, meaning that no iterations of the evolutionary algorithm are executed.}
\label{fig:toggle}
\end{figure}

Next, there is the earlier mentioned restriction on activating layers as shown in the second clause on line 8.
LTPOT has a selection process in place for which pipelines will be evaluated on the entire dataset.
This means that at higher layers, the pipelines in the population are likely already quite good.
Because of these two factors, we want to limit the exploration in higher layers.
To do this, instead of running the evolutionary algorithm in each layer every generation, higher layers can be turned off for some generations.
In this study, for a layered structure with $M$ layers, layer $l$ is progressed for $\min(2^{(M-l+1)}, g)$ generations every $g$ generations, with $l \in \{1,\cdots,M\}$.
This is demonstrated with $g=12$ and $M=4$ in Fig.\ref{fig:toggle}, and checked in the second clause on line 8.
We have not yet evaluated if this leads to significant improvements.\\

Finally, to prevent any one pipeline from halting the algorithm, a pipeline's evaluation is automatically stopped if it exceeds a given time limit.
If the evaluation is stopped this way, the pipeline is marked as failure and will not be considered as parent for the next generation.
This behavior is present in the original TPOT, and adopted to LTPOT by further decreasing the time limit by layer.
In the top layer, each individual is allowed the same evaluation time as it would in TPOT.
However, for lower layers, the time allowed is decreased quadratically proportional to the sample size (a layer with half the data gets a fourth of the time per individual).

\section{Empirical evaluation}\label{sec:results}

\subsection{Experimental Questions}
The goal of LTPOT is to find pipelines at least as good as TPOT's, but in less time. This also means that, given the same amount of time, LTPOT could very well find better pipelines.
To assess whether or not this is achieved, we will evaluate LTPOT in three ways.

First, we want to evaluate if, given the same amount of time, LTPOT will outperform TPOT when their best found pipelines are compared.
To do this, a ranking is constructed between TPOT and LTPOT for each dataset over time, by ranking the performance of the best pipeline found so far at regular time intervals.
We omit a comparison to Random Search, as TPOT compared favorably to Random Search in earlier work~\cite{DBLP:journals/corr/OlsonBUM16}.

Secondly, to quantify how much faster LTPOT is, we compare the time needed for LTPOT to find a pipeline at least as good as the best pipeline found by TPOT.
We then compare it to the time TPOT needs to find this pipeline.

Finally, we will also compare the Area Under the Receiver Operating Characteristic curve (AUROC) of the final pipelines found by each method, so we can quantify the difference in model quality between the methods.

\subsection{Experimental setup}
We compare LTPOT to the original TPOT by evaluating both on a selection of 18 large datasets.
We specifically chose larger datasets so that there will be a distinct difference in time needed to evaluate individuals on the entire dataset versus just a subset.
All datasets in the selection contain at least one hundred thousand instances, though most contain exactly one million. The selection includes pseudo-artificial datasets described in~\cite{vanRijn2014BNG}. 
The datasets are available for download and inspection on OpenML\footnote{https://www.openml.org/s/69}, an open database for machine learning experiments~\cite{OpenML2013}.
We previously evaluated LTPOT on  a selection of datasets from the Penn Machine Learning Benchmark~\cite{DBLP:journals/corr/OlsonCOUM17}, which TPOT was initially evaluated on.
Because those datasets were relatively small, pipeline evaluations were quick even on the full dataset, so there was no significant benefit of using LTPOT.
On each dataset, each method is evaluated nine times, starting with a different initial generation each time.\\

As described earlier, there are many hyperparameters with which to tune LTPOT.
In this study, we only experiment with $g$, the amount of generations between transfer.
The choices for $g$ will be $2$ and $16$, and these configurations will be referred to as LTPOT-2 and LTPOT-16, respectively.
This is meant to give insight in the effectiveness of two functions of LTPOT: filtering and early optimization.
For LTPOT-2, layers act almost solely as a filter, by passing the best individuals to the next layer every other iteration, not much optimization takes place in lower layers, but it does allow for the early discarding of pipelines which seem to perform poorly.
With LTPOT-16, we instead see that a lot of optimization can take place based on results found in lower layers, as relatively more time is spent evaluating and improving individuals in lower layers compared to LTPOT-2.
In other words, LTPOT-2's first layer allows for more early exploration, while LTPOT-16's first layer is more focused on exploitation.\\

The amount of individuals transferred, $k$, will be set to $15$, which is half of the total population size $P=30$.
Each LTPOT configuration, as well as TPOT, is run nine times per dataset, each time with a different \emph{random seed}, guaranteeing a different initial population and subsequent choices for crossover and mutation. Each run set to last 8 hours, but each individual pipeline may only be evaluated for at most 10 minutes.
We explored different values for $P$ and different amounts of evaluation time per individual, while keeping the total run time constant at 8 hours, and found that for the chosen datasets these values strike a balance between having a diverse enough population and being able to evaluate enough generations.

\subsection{Results}
\begin{figure}
\centering
\includegraphics[width=1\linewidth]{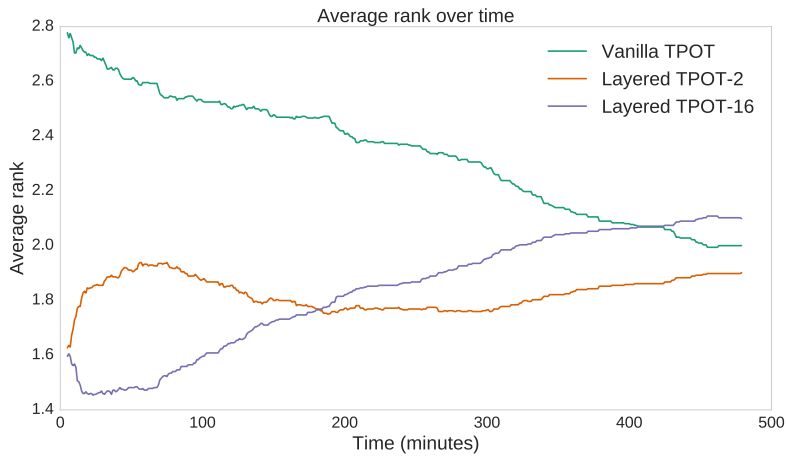}
\caption{Ranking of each method averaged over all datasets based on internal AUC scores of the best individual found so far.}
\label{fig:tpotranking}
\end{figure}

First, we compare the various configurations by their average rank over time, which can be seen in Fig.~\ref{fig:tpotranking}.
In this figure, for each configuration, for every dataset and seed, the best found pipelines so far are ranked against each other every minute.
For each method, the average Friedman rank~\cite{Demsar:2006p14241} across all these datasets and seeds is calculated based on the highest AUROC score achieved by the best pipeline so far, using the fixed hyperparameter values stated above.
To calculate the rank, we consider the result a tie if the difference in AUROC values is smaller than 0.1. A lower rank is better.

In Fig.~\ref{fig:tpotranking}, we see that LTPOT on average achieves the best scores throughout the entire 8 hour period. LTPOT-16 starts by outperforming LTPOT-2 slightly, but as time goes on its relative performance drops, even being surpassed by TPOT.
From this, it seems that while LTPOT works well as a filter, optimization in early layers does not pay off beyond the early stages.
Thus, a lower value for $g$ is better based on these results.
Furthermore, we see that towards the end TPOT is decisively improving over LTPOT-16, but only very slowly over LTPOT-2, as the average rank of LTPOT-2 increases only very slowly.

A different rank does not necessarily mean the found model performance differs a lot. To clarify this, we look at the difference in AUROC by dataset per configuration, as seen for 9 of 18 datasets in Fig.~\ref{fig:auc}.\footnote{A similar figure for the other 9 datasets can be found in Appendix A.}

In Fig.~\ref{fig:auc} we show a boxplot that describes the distribution of AUROC scores of the final pipelines by each method. 
No single method is dominant over the others, and differences in AUROC scores are small for almost all datasets. Using a student t-test, we determined that there is no statistically significant difference ($p<0.05$) between the final pipelines  (after the full 8 hour time budget).

\begin{figure}
\centering
\includegraphics[scale=0.47]{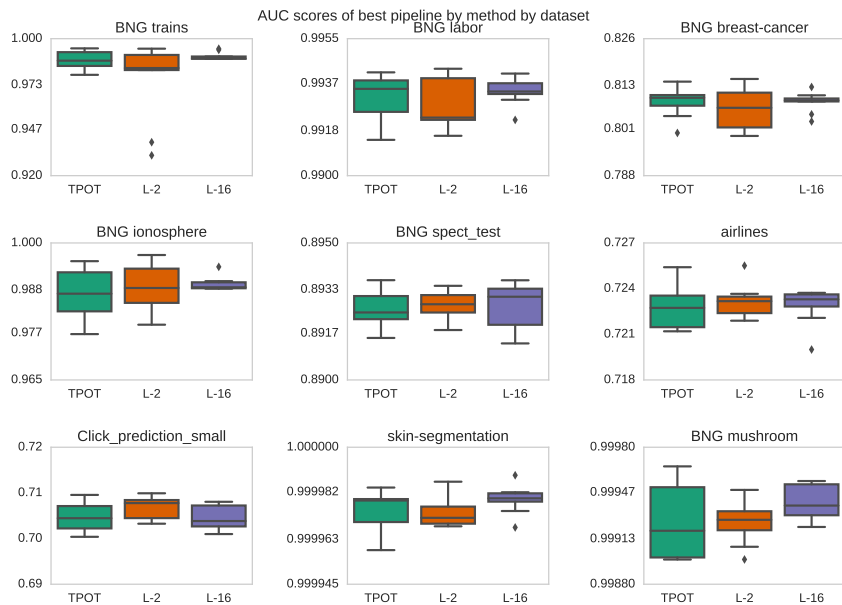}
\vspace{0cm}
\caption{An overview of achieved AUROC score for each run of each method by dataset.}
\label{fig:auc}
\end{figure}

\begin{figure}
\centering
\hspace*{-1.3cm}
\includegraphics[scale=0.25]{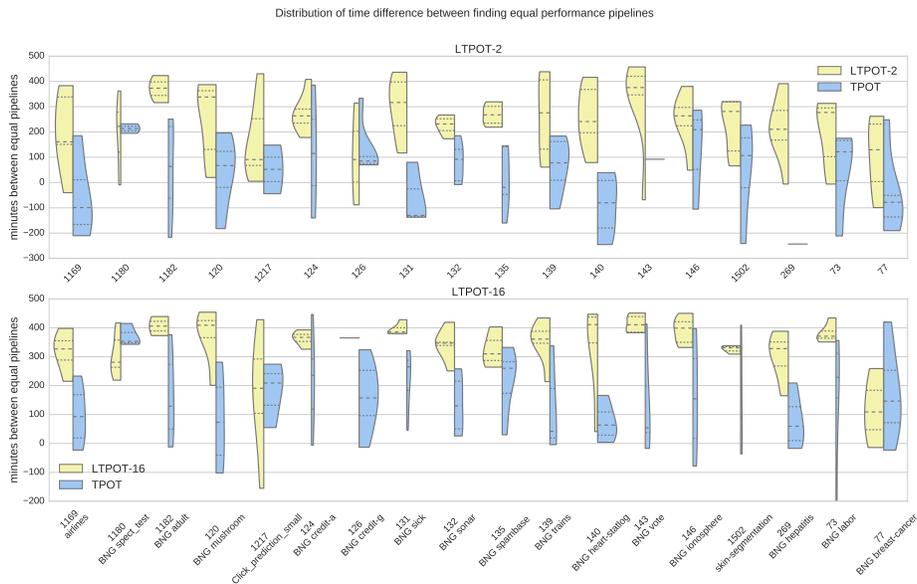}
\vspace{0cm}
\caption{Violin plots of the time difference between finding two equally good pipelines. Datasets are shown together with their OpenML ID number.}
\label{fig:violin}
\end{figure}

However, looking at the average ranking by configuration in Fig.~\ref{fig:tpotranking}, we see that under smaller time budgets LTPOT-2 often finds pipelines which are better than TPOT.
In this scenario, it is interesting to see how much time LTPOT needs to find a pipeline at least as good as the best pipeline TPOT found.
We compared LTPOT-2 to TPOT for each dataset and seed, and looked at how long it took for the method which found the best pipeline to find a pipeline at least as good as the best found pipeline by the other method. Figure~\ref{fig:violin} shows the time difference (in minutes) between finding these equally good pipelines. Positive values indicate that LTPOT is faster.
Yellow distributions correspond to seeds where LTPOT eventually found the best pipeline, and show how much sooner LTPOT found a pipeline at least as good as the best pipeline of TPOT. Blue distributions correspond to cases where TPOT eventually found the best pipeline, and show how much later TPOT found a pipeline at least as good as LTPOT's best.

In general, when LTPOT finds the best pipeline, it finds a pipeline at least as good as TPOT's best pipeline much sooner. In particular for LTPOT-16 we see that it often is at least 200 minutes faster.
Even in the cases where TPOT eventually finds the best pipelines, we see that it often finds a pipeline at least as good as LTPOT's best only after LTPOT already found it. This again is especially true for LTPOT-16, where almost all yellow distributions are entirely positive.\\

\begin{figure}
\centering
\hspace*{-1.4cm}
\includegraphics[scale=0.40]{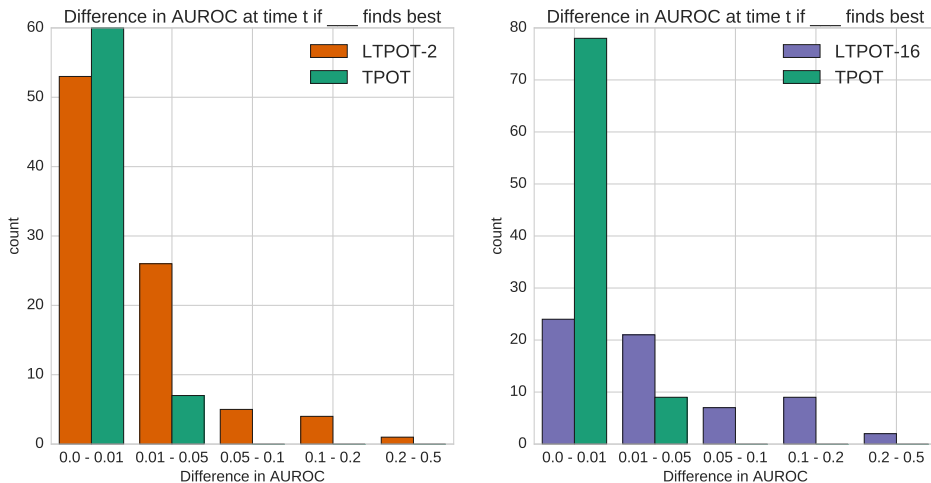}
\caption{Shows the AUROC difference at time $t$, which is the time the best method (color coded) finds a pipeline at least as good as the other method will find.}
\label{fig:time_t}
\end{figure}

However, even when one method finds a pipeline at least as good as the other method's eventual best pipeline, it can still be the case that the eventual worst method at that same time already has quite a good pipeline.
Therefore, we compare the performance of the best pipeline of each method found at time $t$, where time $t$ is the time where the best method finds a pipeline at least as good as the eventual best pipeline of the worst method. Between 18 datasets and 9 seeds, there are 162 comparisons between TPOT and LTPOT-2 or LTPOT-16. The comparison in AUROC at time $t$ is shown in Figure~\ref{fig:time_t}.

We see that when TPOT finds a pipeline at least as good as LTPOT's best, in most cases it does so when LTPOT already has found a pipeline at most 0.01 AUROC worse. 
However, when LTPOT finds a pipeline at least as good as TPOT's best, TPOT has relatively worse pipelines more often, and in some cases as much as over 0.2 AUROC worse.

From this we conclude that in many cases LTPOT finds good pipelines faster. While LTPOT-2 does not always find the best pipeline, when it doesn't, it finds comparable pipelines at least as quickly as TPOT. LTPOT-16 finds comparable pipelines even quicker, although it becomes less competitive under larger time budgets.

\subsection{Future work}
The structure of Layered TPOT allows for more hyperparameters to be tuned, and their effect remains unexplored at this point.
The number of layers as well as the effect of their granularity could possibly be tuned to the dataset.
Whether or not to turn off higher layers, and with which frequency, should be explored as well.
The amount of individuals to transfer between layers may also change how quickly higher layers can optimize pipelines.
The choice of the amount of generations before transfer, $g$, will influence how much optimization will be done in the lower layers.
Perhaps these parameters should change over time in a single run.
All of these choices come in addition to the hyperparameters already available for TPOT, such as the population size, mutation rate and crossover rate.

Finally, there are possible changes not yet captured in hyperparameters.
For instance, it might be better to favor crossover over mutation in higher layers, so that the focus of exploration shifts to combining promising pipelines in higher layers.
It could be that having a big population in the higher layers is unnecessary, and shrinking population size there might yield similar results in a faster time frame.
Currently, the way to evaluate individuals quickly is to sample a number of instances of the data.
Instead, one could create a subset by selecting features, or apply compression techniques to represent the data. We will explore these aspects in subsequent work.

Results of this study will be made available on OpenML\footnote{www.openml.org}, and the code for Layered TPOT can be found on Github\footnote{https://github.com/PG-TUe/tpot/tree/layered}.

\section{Conclusion}\label{sec:conclusion}
In this paper we presented an extension to TPOT called Layered TPOT.
In the extension, instead of evaluating each pipeline on all data, only pipelines which have shown good results on subsets of the data are evaluated on all data.
It does this by introducing layers. In each layer a distinct group of individuals is subject to the evolutionary algorithm, but individuals are only evaluated on a subset of the data as defined by the layer.
Each subsequent layer will evaluate the individuals on more data, and if an individual performs well in one layer, it will be transferred to the next, to be trained on more data and compete with other promising pipelines trained on the same subset size.

To determine the usefulness of LTPOT, two configurations have been compared to TPOT, on a selection of 18 large datasets.
The results showed that LTPOT is not strictly better than TPOT, but it often finds good pipelines much faster, and under smaller time budgets, it outperforms TPOT on average. 
Moreover, LTPOT allows for a lot of flexibility in the configuration of its structure, and the effects of changes to these configurations remain to be explored.

\bibliographystyle{splncs}
\bibliography{references}

\clearpage

\appendix
\section{Additional Experiment Results}

Below is a figure similar to Figure\ref{fig:auc} for the other 9 datasets in the benchmark.

\begin{figure}
\centering
\includegraphics[scale=0.47]{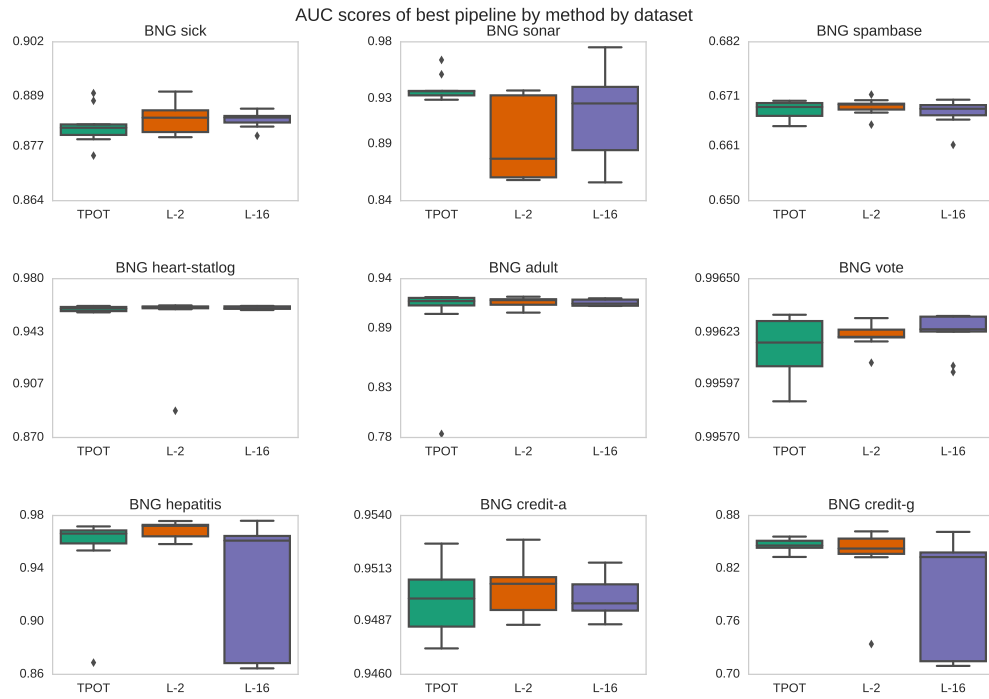}
\vspace{0cm}
\caption{An overview of achieved AUROC score for each run of each method by dataset.}
\label{fig:auc2}
\end{figure}

\begin{figure}
\centering
\hspace*{-1.4cm}
\includegraphics[scale=0.40]{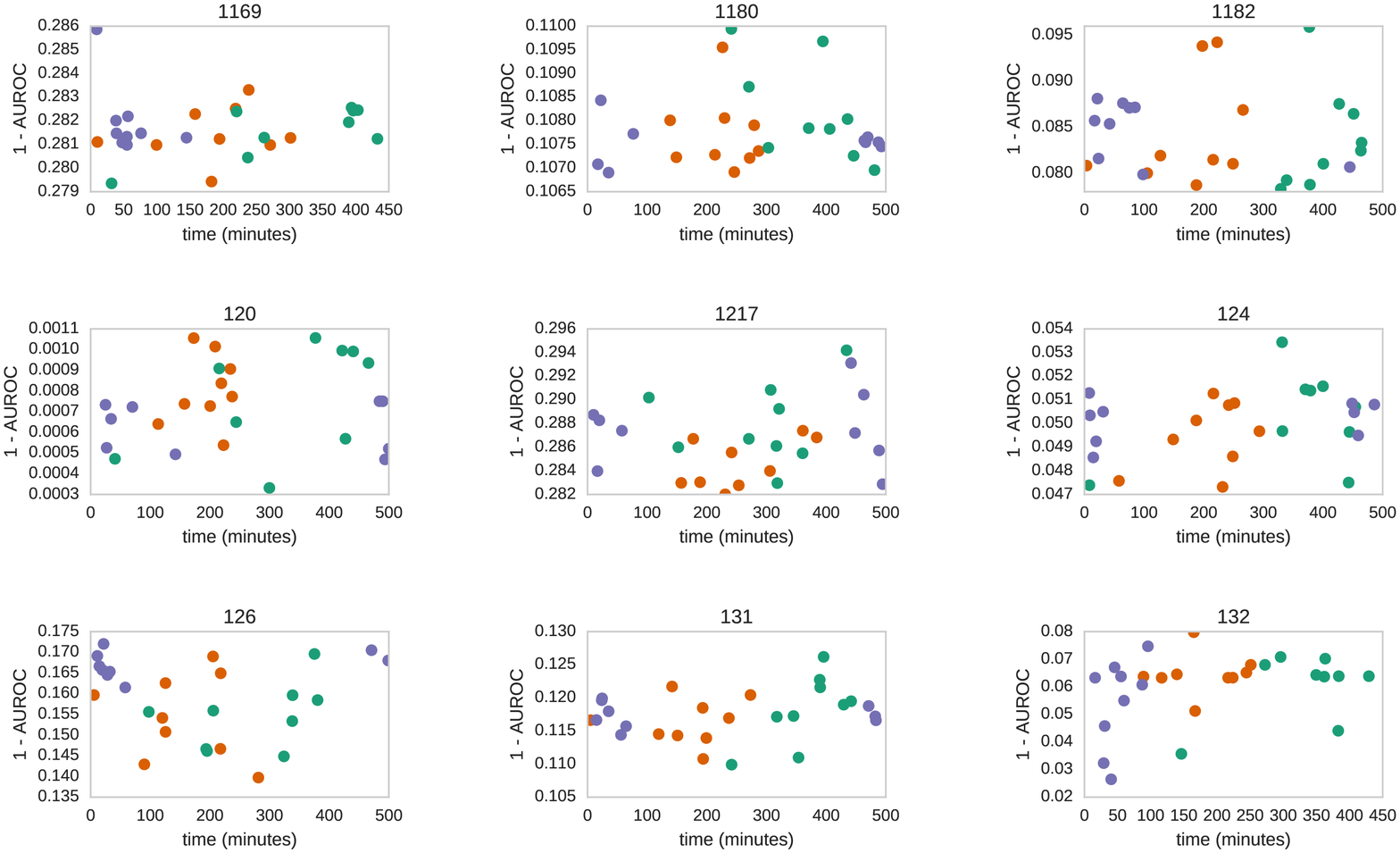}
\caption{Shows per dataset per seed per method when the best pipeline was found, and its '1-AUROC' score.}
\label{fig:end_1}
\end{figure}

\begin{figure}
\centering
\hspace*{-1.4cm}
\includegraphics[scale=0.40]{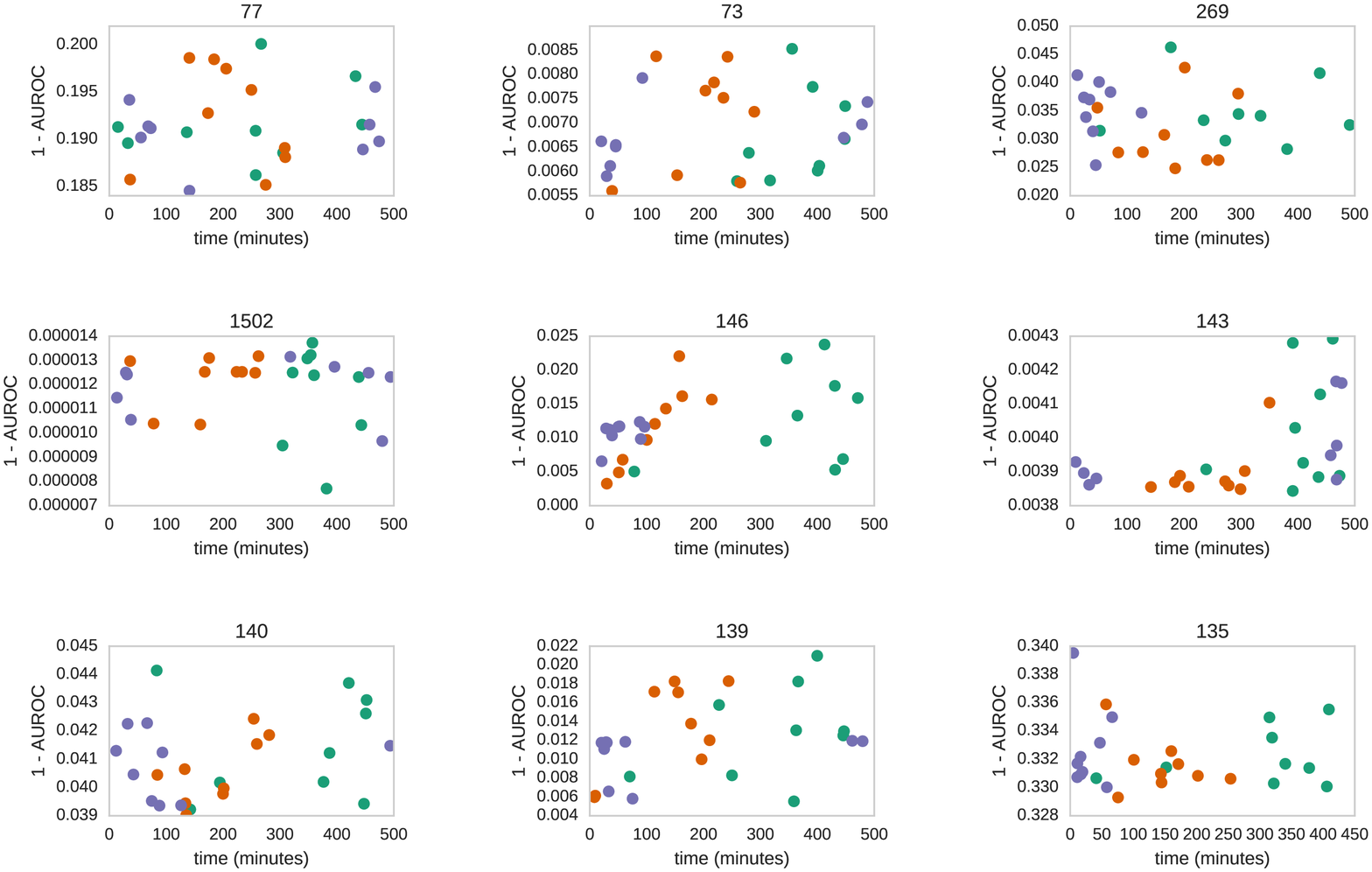}
\caption{Shows per dataset per seed per method when the best pipeline was found, and its '1-AUROC' score.}
\label{fig:end_2}
\end{figure}

Figures~\ref{fig:end_1} and~\ref{fig:end_2} show the score of the best pipeline and what time it was found per dataset, method and seed.

\end{document}